\pdfoutput=1

\documentclass[11pt]{article}
\usepackage{authblk}

\usepackage{ACL2023}

\usepackage{times}
\usepackage{latexsym}
\usepackage{amsmath}
\usepackage{svg}

\usepackage{graphicx}
\usepackage{subcaption}
\usepackage{makecell}
\usepackage{multirow}
\usepackage{algorithm}
\usepackage{algorithmic}
\usepackage{amsmath}
\usepackage{amsfonts}
\usepackage{booktabs}
\usepackage{siunitx}
\usepackage{float} 
\usepackage{enumitem} 
\usepackage{booktabs}
\usepackage{listings}
\usepackage{soul}
\usepackage{xcolor}
\usepackage{enumitem}

\definecolor{lightblue}{RGB}{207,226,245}

\usepackage[T1]{fontenc}

\usepackage[utf8]{inputenc}

\usepackage{microtype}

\usepackage{inconsolata}
\usepackage{tabularx}
\newcommand{\CorpusName}{RAGTruth}
\NewDocumentCommand{\shizhe}
{ mO{} }{\textcolor{blue}{\textsuperscript{\textit{shizhe}}\textsf{\textbf{\small[#1]}}}}

\title{
\CorpusName: 
A Hallucination Corpus for Developing Trustworthy Retrieval-Augmented Language Models
}

\author[1]{\textbf{Cheng Niu}}
\author[1]{\textbf{Yuanhao Wu}}
\author[1]{\textbf{Juno Zhu}}
\author[1]{\textbf{Siliang Xu}}
\author[1]{\textbf{Kashun Shum}}
\author[1]{\\\textbf{Randy Zhong}}
\author[1]{\textbf{Juntong Song}}
\author[2]{\textbf{Tong Zhang}}
\affil[1]{NewsBreak}
\affil[2]{University of Illinois Urbana-Champaign}
\affil[ ]{ {cheng.niu@newsbreak.com}}

\begin{document}
\maketitle
\begin{abstract}

Retrieval-augmented generation (RAG) has become a main technique for alleviating hallucinations in large language models (LLMs). 
Despite the integration of RAG, LLMs may still present unsupported or contradictory claims to the retrieved contents. 
In order to develop effective hallucination prevention strategies under RAG, it is important to create benchmark datasets that can measure the extent of hallucination. 
This paper presents {\CorpusName}, a corpus tailored for analyzing word-level hallucinations in various domains and tasks within the standard RAG frameworks for LLM applications. 
{\CorpusName} comprises nearly 18,000 naturally generated responses from diverse LLMs using RAG. 
These responses have undergone meticulous manual annotations at both the individual case and word levels, incorporating evaluations of hallucination intensity. 
We not only benchmark hallucination frequencies across different LLMs, but also critically assess the effectiveness of several existing hallucination detection methodologies. 
We show that using a high-quality dataset such as {\CorpusName}, it is possible to finetune a relatively small LLM and achieve a competitive hallucination detection performance when compared to the existing prompt-based approaches using state-of-the-art LLMs such as GPT-4. Furthermore, the finetuned model can effectively mitigate hallucination in LLM responses.
\end{abstract}

\section{Introduction}
Large language models (LLMs) have achieved remarkable success in a variety of tasks, including text generation~\cite{li2022pretrained}, machine translation~\cite{kocmi2023large}, and question answering~\cite{zhao2023survey}. However, one of the key challenges in deploying LLMs in real-world applications is their tendency to hallucinate~\cite{kaddour2023challenges}.
Hallucination in the context of LLMs usually refers to a situation where the model generates content that is not based on factual or accurate information~\cite{Rawte2023}. 
The occasional generation of outputs that appear plausible but are factually incorrect significantly undermine the reliability of LLMs in real-world scenarios, such as medical diagnoses~\cite{pal2023medhalt} and news summarization~\cite{shen2023why}.

\begin{table}[t]

\renewcommand{\arraystretch}{1.2}

\small
\centering
    \begin{tabular}{p{0.45\textwidth}}
    \toprule
    \textbf{\textsc{Question:}} how to prepare to get an ultrasound? \\
    \sethlcolor{orange!20}
    \textbf{\textsc{Reference:}} Passage 1: 1 If you are just interested in checking the gender, you can go as early as 17 weeks. 2  Any earlier, and the results may not be accurate, leading to an extra coat of paint for your bedroom. 3 To get good pictures, 3D ultrasounds are best performed between 20 and 32 weeks, and \hl{the best pictures are between 24 and 30 weeks}. \\ ...\\
    \textbf{\textsc{Response:}} Based on the given passages, here's how to prepare for an ultrasound:\\
    * For a gender reveal ultrasound, wait until at least 17 weeks of pregnancy to ensure accurate results. \\
    \sethlcolor{lightblue}
    * For 3D ultrasounds, schedule the appointment \hl{between 20 and 32 weeks of pregnancy for the best pictures} ...\\
    \midrule
    \textbf{\textsc{Annotation}}\\
    \textbf{Span}: between 20 and 32 weeks of pregnancy for the best pictures\\
    \textbf{Type}: Evident Conflict\\
    \textbf{Reason}: Original: "the best pictures are between 24 and 30 weeks", Generative: "between 20 and 32 weeks of pregnancy for the best pictures"\\
    \bottomrule
    \end{tabular}
    \caption{An example of RAGTruth data from the question answering task. It contains context, response generated by LLM with and span-level annotation.}
    \label{tab:1-dataset-example}
\end{table}

To reduce hallucination, various methods have been developed that can be applied at different stages of LLM lifecycle, including pre-training~\cite{brown2020language}, supervised fine-tuning~\cite{zhou2023lima,zhang2023r}, RLHF~\cite{ouyang2022training, lin-etal-2022-truthfulqa}, and inference~\cite{dhuliawala2023chainofverification, gao-etal-2023-rarr}. 
In terms of detection, methods are developed by examining the model's intrinsic state~\cite{guo2017calibration}, comparing it with external data and tools~\cite{chern2023factool}, or leveraging the LLM's inherent powerful capabilities for self-checking~\cite{agrawal2023language,Manakul2023-selfcheckgpt}. 
Retrieval-augmented generation (RAG) is extensively used to supply LLMs with updated, relevant knowledge, significantly mitigating hallucination~\cite{varshney2023stitch}. 
Nevertheless, even with RAG and other enhancements, LLMs still produce statements that are either unfounded or contradict the information provided in the retrieved references~\cite{shuster2021retrieval}.

Despite the growing awareness of the hallucination phenomenon, the understanding of hallucination in LLMs is still in its early stages. 
One key challenge is the lack of high-quality, large-scale datasets specifically designed for hallucination detection. 
This issue is particularly acute in RAG settings. 
Due to the relatively low hallucination ratio, a substantial increase in annotation resources is needed.
Existing datasets for LLM hallucination detection are predominantly synthesized~\cite{Li2023}. 
For instance, in~\citet{memotrap2023,longpre-etal-2021entity}, prompts conflicting with conventional knowledge are purposely generated to trigger hallucinations. 
While these approaches are efficient at generating hallucinations, the resulting artificial hallucinations can substantially differ from those that naturally occur.
In~\citet{chen2023felm, hu-etal-2023-bschecker}, hallucination datasets are developed by manual annotations of naturally produced LLM responses. However, these datasets are of limited size and are not specifically focused on the RAG scenario.

In this paper, we introduce a large-scale high-quality dataset specifically designed for word-level hallucination detection for RAG applications. 
Using this dataset, we have conducted an extensive benchmarking of mainstream LLMs to assess their tendency to generate hallucinations, as well as evaluate current methods for hallucination detection. 
Additionally, we have demonstrated superior performance in identifying hallucinations by fine-tuning LLM with {\CorpusName} dataset. 
Our key contributions are:
\begin{enumerate}[label=(\roman*)]
\item We propose {\CorpusName}, a large-scale word-level hallucination evaluation dataset specifically for the RAG scenario across several common tasks. 
It consists of nearly 18,000 fully annotated natural responses generated from major open-source and closed-source LLMs. 
\item We perform a comprehensive comparison of different hallucination detection methods at both the passage and word levels.
\item We present a baseline method of fine-tuning LLM for hallucination detection. 
It is shown that by fine-tuning the Llama-2-13B model on the {\CorpusName} training data, we can achieve results competitive to the existing prompt-based approaches using GPT-4. 
This shows the potential of developing better hallucination detection methods using {\CorpusName}.
\item We show that by using our finetuned hallucination detector, it is possible to significantly reduce the occurrence of hallucinations in the responses from LLMs. The improvement holds even for models with inherently low hallucination rates, such as GPT-4.
\end{enumerate}

\section{Related Work}
\subsection{Hallucination of Large Language Models}
Though hallucination in traditional natural language generation (NLG) contexts has been widely studied\cite{10.1145/3571730}, comprehending and tackling this problem in the context of LLMs presents distinct challenges\cite{Zhang2023}.
Existing research has demonstrated that incorporating up-to-date, relevant knowledge in the prompt can effectively reduce fact-conflicting hallucination~\cite{vu2023freshllms, lewis2021retrievalaugmented}. 
This approach,  referred to as {\em Retrieval-Augmented Generation} (RAG), is widely used in real-world LLM applications. 
For instance, Google Bard~\footnote{\url{https://bard.google.com}} and Microsoft BingChat~\footnote{\url{https://www.bing.com}} have implemented this technique. 

\subsection{Hallucination Evaluation Datasets}

Extensive research has focused on hallucination benchmarks within conventional Natural Language Generation settings \cite{dziri2022evaluating, zhong-etal-2021-qmsum, durmus-etal-2020-feqa, lin-etal-2022-truthfulqa}. With the rise of LLMs, the detection of hallucinations has become increasingly challenging, necessitating the development of high-quality datasets for LLM evaluation \cite{chen2023can}. Contributions in this domain include HaluEval \cite{Li2023}, which introduced datasets encompassing both synthetically and naturally generated LLM responses, and FELM \cite{chen2023felm}, which concentrated on naturally generated LLM responses across multiple domain tasks. RefChecker \cite{hu-etal-2023-bschecker}, a distinctive approach, breaks down claims in LLM responses into triples and utilizes human annotation to assess the truthfulness of facts. 
Notably, these works primarily focus on annotating factual hallucinations in LLM responses. Distinguishing from previous research, our work centers on the evaluation of LLMs within RAG settings. 

\begin{figure*}[!ht]
    \centering
    
    \includegraphics[width=\textwidth]{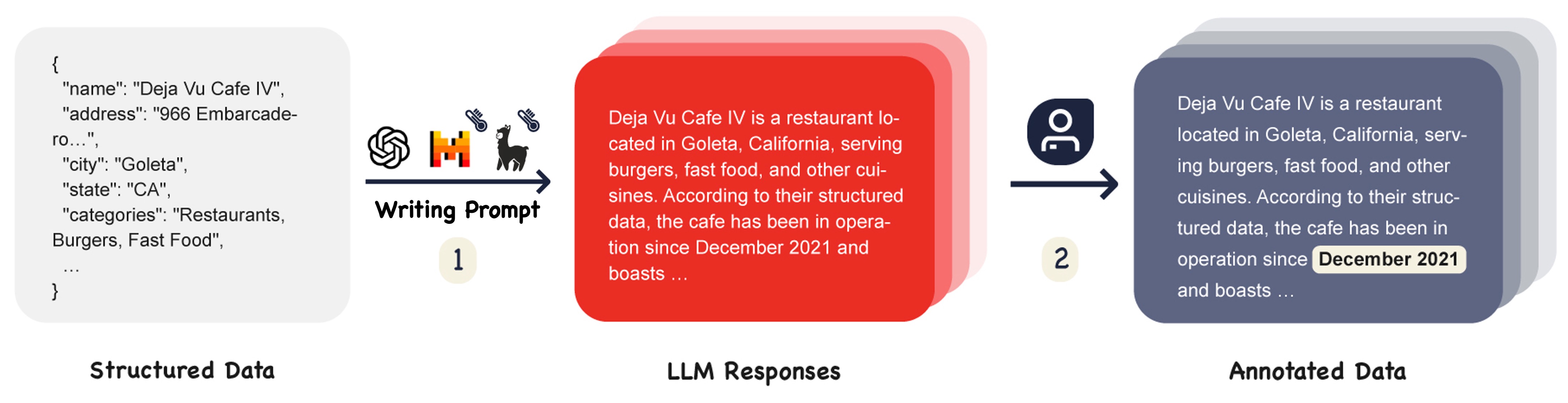}
    \caption{Data gathering pipeline. 
    Taking a data-to-text writing task as an example, our data gathering pipeline includes 2 steps: 
    1) response generation. 
    We generated responses with multiple LLMs and natural prompts. 
    2) human annotation. Human labeler annotated hallucinated spans in LLM responses.}
    \label{fig:dataset}

\end{figure*}

\subsection{Hallucination Detection Methods}
Researchers have been exploring various methods to enhance the reliability of LLMs by detecting hallucinations. 
In \citet{Azaria2023,xiao2021hallucination,malinin2021uncertainty}, intrinsic model uncertainty metrics such as token-level probability and entropy are used to detect hallucinations. 
When direct access to output uncertainty is not feasible, as in the case with limited APIs like GPT-4, an alternative approach involves employing a fully accessible LLM as a proxy~\cite{Manakul2023-selfcheckgpt}.
In \citet{falke2019ranking, barrantes2020nli}, natural language inference modules are adapted to check the information consistency between the articles and their summaries, and it has been shown that external knowledge is helpful for detecting factual hallucinations.~\cite{guo2022survey, Mallen2022}. 
Additionally, methods that leverage the inherent capabilities of LLMs have been proposed for self-checking, such as verbalization-based and consistency-based methods~\cite{xiong2023llms, Manakul2023-selfcheckgpt}. 
These techniques aim to detect hallucinations without relying on internal states or external data and tools.

\section{Construction Process of RAGTruth}
We established a data generation and annotation pipeline as shown in Figure~\ref{fig:dataset}. 

\subsection{Hallucination Taxonomy}
Different from open-end generation, under RAG setting, the prompt contains rich context information, and the model is generally required to generate text based on the provided context. The detection and mitigation of inconsistencies between retrieved information and responses emerge as significant sources of hallucination.

As outlined below, we categorize the hallucination in the RAG setting into four types. For concrete examples of each type, please refer to Appendix~\ref{sec:example}.

\paragraph{Evident Conflict:} for when generative content presents direct contraction or opposition to the provided information. These conflicts are easily verifiable without extensive context, often involving clear factual errors, misspelled names, incorrect numbers, etc.

\paragraph{Subtle Conflict:} for when generative content presents a departure or divergence from the provided information, altering the intended contextual meaning. These conflicts often involve substitution of terms that carry different implications or severity, requiring a deeper understanding of their contextual applications. 

\paragraph{Evident Introduction of Baseless Information:} for when generated content includes information not substantiated in the provided information. It involves the creation of hypothetical, fabricated, or hallucinatory details lacking evidence or support. 

\paragraph{Subtle Introduction of Baseless Information:} is when generated content extends beyond the provided information by incorporating inferred details, insights, or sentiments. This additional information lacks verifiability and might include subjective assumptions or commonly observed norms rather than explicit facts. 

\subsection{Response Generation}

\paragraph{Tasks and Data Sources} 
We selected three widely recognized generation tasks with RAG settings for response generation: Question Answering, Data-to-text Writing, and News Summarization.

For the task of question answering, we conducted a random sampling from the training set of MS MARCO~\cite{DBLP:journals/corr/NguyenRSGTMD16}. 
To reduce the difficulty of annotation, we selected only those questions related to daily life, and preserved only three retrieved passages for each question.
Then we prompted LLMs to generate answers for each question solely based on the retrieved passages.

For the data-to-text writing task, we prompted LLMs to generate an objective overview for a randomly sampled business in the restaurant and nightlife categories from the Yelp Open Dataset~\cite{yelp2023}.  
In this dataset, information pertaining to a business is represented using structured data.
To streamline the annotation process, we focused only on the following business information fields: \textit{BusinessParking}, \textit{RestaurantsReservations}, \textit{OutdoorSeating}, \textit{WiFi}, \textit{RestaurantsTakeOut}, \textit{RestaurantsGoodForGroups}, \textit{Music}, and \textit{Ambience}. 
In addition to the structured data, we have also included up to three business-related user reviews to enrich the context information.
In the prompt, these information is represented in JSON format.

For the news summarization task, we randomly selected documents from the training set of the well-known CNN/Daily Mail dataset~\cite{see-etal-2017get} as well as recent news articles from a prestigious news platform. 
LLMs were prompted to generate a summary for each of the source news.  

\paragraph{Models} 
The following six models with strong instruction-following ability are used for response generation: GPT-3.5-turbo-0613 and GPT-4-0613 from OpenAI~\cite{openai2023gpt4}; Mistral-7b-Instruct from Mistral AI~\cite{jiang2023mistral}; Llama-2-7B-chat, Llama-2-13B-chat and Llama-2-70B-chat (4bit quantized)\footnote{\url{https://huggingface.co/TheBloke/Llama-2-70B-Chat-AWQ}} from Meta~\cite{touvron2023llama}. To ensure a fair comparison, the prompts used for response generation are kept straightforward with subtle differences among various models to optimize their performance. We provide detailed prompts in the Appendix~\ref{sec:prompts}. 

For each sample, we collected one response from each model.
As a result, we got a total of 6 responses for each input sample.



\subsection{Human Annotation}

Identifying AI-generated hallucinations is a challenging task. It requires a strong capacity for critical thinking to understand the logical flow of various texts, along with meticulous attention to detail for spotting subtle inaccuracies and inconsistencies. 
Moreover, a certain level of media literacy and knowledge of current affairs is crucial to grasp the subjects discussed in news-related sample data. 
Therefore, we chose annotators who are proficient in English and possess a bachelor's degree in English, Communications, or relevant fields to ensure the accuracy and reliability of the annotation results. We recruited annotators from a professional vendor and paid them at a rate of \$25 per hour per individual.

The annotators are invited to perform annotation tasks using Label Studio~\cite{labelstudio2023}. Each labeling task is presented within one page, comprising the following components: 1) the context provided to the AI models; 2) a set of 6 responses, generated by different AI models. Our annotation interface is available in Appendix \ref{sec:interface}.

Their task was to annotate the specific spans of the generated text that contains hallucinated information and categorize them into the four types.
To ensure the quality of the annotations, each response is independently labeled by two annotators. 
The consistency rate of two annotators was 91.8\% at the response level and 78.8\% at the span level.
In cases where there is a considerable difference between the two annotations, a third review is undertaken. 

\subsection{Annotations for Adaptive Evaluation}
In different contexts, the definition and criteria for hallucination vary, and the annotation of hallucination is not always straightforward. In contentious cases, additional annotations are provided to accurately reflect these situations. This approach enables users to adopt various evaluation strategies tailored to their specific application circumstances. Please refer to Appendix \ref{sec:interface} for more statistical information about these annotations.


\paragraph{Implicit Truth} The extensive world knowledge and ability of LLMs is a significant advantage in open-ended generation scenarios. But in the context of this paper, which focuses on the relatively strict RAG scenarios, we have labeled information that is not mentioned in the reference but may be truthful as hallucinations. 
For instance, mentioning a local officer's name not present in the reference or claiming that a restaurant accepts credit card payments without any basis.

The decision is based on the observation that LLMs have a relatively high chance of making errors when generating detailed facts, partly because their embedded knowledge can be outdated. Therefore, RAG applications usually instruct LLMs not to generate factual content without the support of references.
Besides, we provided an additional span-level annotation named \textit{implicit\_true} for these spans to accommodate different application needs. 


\begin{table*}[t]
\centering
\scriptsize
\begin{tabular}{l|ccccccccc}
\toprule
\multirow{2}{*}{Task} & \multirow{2}{*}{\# Instance} & \multirow{2}{*}{\# Resp.} & \multicolumn{2}{c}{\textsc{Context Length}} & \multicolumn{2}{c}{\textsc{Resp. Length}} &\multicolumn{3}{c}{\textsc{Hallucination}} \\
\cmidrule(lr){4-5} \cmidrule(lr){6-7} \cmidrule(lr){8-10}
  & & & Mean  & Max  & Mean  & Max & \# Resp. & \% Resp. & \# Span \\
\midrule
Question Answering & 989 & 5934 & 243 & 509 & 119 &381 & 1724 & 29.1\% & 2927 \\
Data-to-text Writing & 1033  & 6198 & 354 & 1253 & 159 & 369 & 4254 & 68.6\% & 9290 \\
Summarization(CNN/DM) & 628 & 3768 & 648 & 1749 & 124 & 632 & 1165 & 30.9\% & 1474 \\
Summarization(Recent News) & 315 & 1890 & 369 & 481  & 89  & 240  & 521 & 27.6\% & 598 \\
\midrule
Overall & 2965 & 17790 & 381 & 1749 & 131 & 632 & 7664 & 43.1\% & 14289 \\
\bottomrule
\end{tabular}
\caption{The basic statistics of \CorpusName. Here "Resp." stands for "Response".}
\label{tab:basic_stat}
\end{table*}

\paragraph{Differences in Handling Null Value} In the data-to-text writing task, certain fields sometimes are with null values. We observed that in the generated results, null is often interpreted as false by some models. Since the more common expressions for negation in our dataset are the boolean value \textit{False} or the text \textit{No}, we labeled these instances as hallucinations (evident introduction of baseless info) and provided a special span-level annotation named \textit{due\_to\_null} for these spans. In the subsequent hallucination detection experiments, our prompts will be aligned with this standard.

\begin{figure}[t]
    \includegraphics[width=0.49\textwidth]{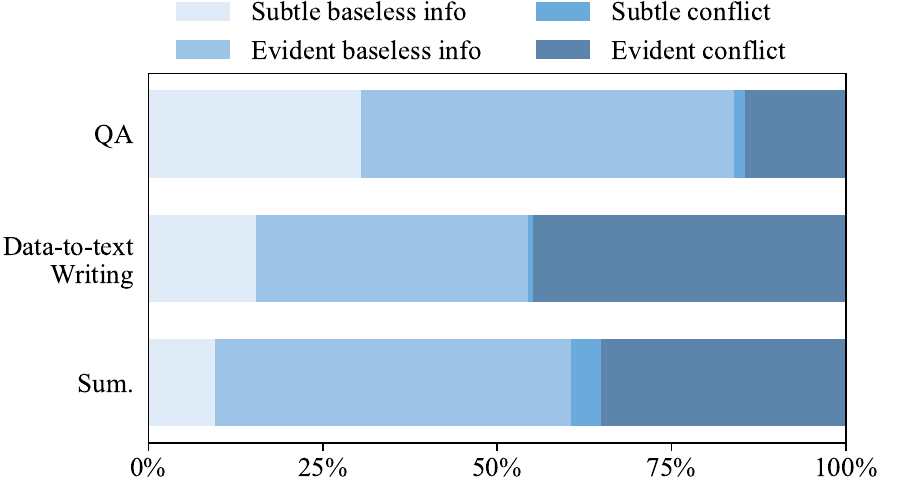}
    \caption{Frequency of different types of hallucination by task.}
    \label{fig:hallu_type}
\end{figure}

\section{Hallucination Benchmark Analysis}

\subsection{Basic Statistics}
We presented detailed statistics of {\CorpusName} in Table~\ref{tab:basic_stat}. Compared to existing datasets for hallucination detection~\cite{Cao2023AutoHallAH, Kamoi2023WiCERE}, the {\CorpusName} dataset is considerably large in scale. The corpus contains a total of 2,965 instances of data, which include 989 instances for question answering, 1,033 instances for date-to-text writing, and 943 instances for news summarization. Each instance comprises responses from 6 different models. 
As shown in Table~\ref{tab:basic_stat}, the {\CorpusName} dataset also features longer prompt and response lengths than existing datasets for hallucination detection~\cite{wang2020asking}. 



\subsection{Hallucination Statistics}

\paragraph{Hallucination Types} As shown in Figure~\ref{fig:hallu_type}, the generation of information baseless in the context was significantly more prevalent than the generation of information conflicting with the context, especially for the question answering tasks. Within the two major categories of \textit{baseless info} and \textit{conflict}, the more severe hallucinations, namely \textit{Evident baseless info} and \textit{Evident conflict}, respectively, account for a significant portion. This observation highlights the importance and challenges of LLMs hallucination mitigation, even in RAG settings.


\paragraph{Hallucination vs Tasks} As shown in Table~\ref{tab:basic_stat}, across the three tasks, the date-to-text writing task exhibited the highest frequency of hallucinations in its responses. 
Inconsistent handling of JSON format data, especially time and attributes, contributed to a significant number of hallucinations in this task. 
Interestingly, the models did not show a higher rate of hallucinations for recent news compared to outdated news. This could be attributed to the shorter context length in the recent news subtask compared to the CNN/DM subtask.


\begin{table*}[!ht]
\centering
\scriptsize
\begin{tabular}{l|ccccccccccc}
\toprule
\multirow{2}{*}{Model}  & \multicolumn{3}{c}{\textsc{Question Answering}} &\multicolumn{3}{c}{\textsc{Data-to-text Writing}} & \multicolumn{3}{c}{\textsc{Summarization}} &\multicolumn{2}{c}{\textsc{Overall}}\\
\cmidrule(lr){2-4} \cmidrule(lr){5-7} \cmidrule(lr){8-10}  \cmidrule(lr){11-12}
    & \# Resp. & \# Span & Density & \# Resp.  & \# Span & Density & \# Resp.  & \# Span & Density & \# Resp.  & \# Span\\
\midrule
GPT-3.5-turbo-0613    & 75 & 89 & 0.12 & 272 & 384 & 0.18 & 54 & 60 & 0.05 & 401  &   533    \\
GPT-4-0613   & 48 &  51 & 0.06 & 290 & 354 &  0.27 &  74 &  80 & 0.08 & 406  & 485    \\
Llama-2-7B-chat  & 510 & 1010 & 0.59 & 888 & 1775 & 1.27 & 434 & 517 & 0.58 & 1832 &  3302   \\
Llama-2-13B-chat    & 399 & 654 & 0.48  & 983 & 2803 & 1.53 & 295 & 342 & 0.41 & 1677 &  3799   \\
Llama-2-70B-chat$^{\dagger}$  & 320 & 529 & 0.40 & 863 & 1834 & 1.15 & 212 & 245 & 0.26  & 1395 & 2608    \\
Mistral-7B-Instruct  & 378 & 594 & 0.59 & 958 & 2140 & 1.51 & 617 & 828 & 0.86 & 1953 & 3562   \\
\bottomrule
\end{tabular}
\caption{Hallucination counts and density of models. 
$\dagger$: We used 4-bit quantized version of Llama-2-70B-chat.}
\label{tab:halu-model}
\end{table*}

\begin{table}[!t]
\scriptsize
\centering
    \begin{tabular}{c|ccc}
    \toprule
    CLB & \textsc{Summarization} & \textsc{D2T Writing} & \textsc{QA}\\
    \midrule
     1 &  $0.29_{(176,368]}$ & $1.51_{(178,273]}$ & $0.50_{(131,187]}$ \\
     2 &  $0.36_{(368,587]}$ & $1.48_{(273,378]}$ & $0.51_{(187,288]}$ \\
     3 &  $0.44_{(587,1422]}$ & $1.49_{(378,731]}$ & $0.49_{(288,400]}$ \\
    \bottomrule
    \end{tabular}
    \begin{tabular}{c|ccc}
    \toprule
    RLB & \textsc{Summarization} & \textsc{D2T Writing} & \textsc{QA}\\
    \midrule
     1 &  $0.34_{(44,87]}$ & $1.20_{(93,131]}$ & $0.21_{(19,93]}$ \\
     2 &  $0.32_{(87,119]}$ & $1.59_{(131,175]}$ & $0.37_{(93,138]}$ \\
     3 &  $0.44_{(119,245]}$ & $1.69_{(175,258]}$ & $0.87_{(138,257]}$ \\
    \bottomrule
    \end{tabular}
    \caption{Average number of hallucinations per response in different context length buckets (CLB) and response length buckets (RLB) for the three types of tasks. The subscript denotes the minimum and maximum length of this bucket.}
    \label{tab:cl-vs-halu}
\end{table}

\paragraph{Hallucination vs Models} Table~\ref{tab:halu-model} illustrates that among the data we collected, OpenAI's two models demonstrated notably lower hallucination rates compared to others. Specifically, GPT-4-0613 exhibited the lowest hallucination frequency. 

To more clearly compare the hallucination rate of different models, we calculated the hallucination density for each model across three tasks. Hallucination density is defined as the average number of hallucination spans per hundred words in the responses. In the Llama2 series, a clear negative correlation was observed between the model scale and hallucination density, aside from the data-to-text writing tasks. Despite its strong performance in various benchmarks and leaderboards~\cite{zheng2023judging}, 
the Mistral-7B-Instruct model generated the highest number of responses containing hallucinations.

\paragraph{Hallucination vs Length} After removing the top and bottom 5\% of outliers, we partitioned the data for each task type into three equal-sized groups according to the length of the context/response. We then computed the average number of hallucinated spans per response within each group. As shown in Table~\ref{tab:cl-vs-halu}, there is a clear overall trend of an increase in the average number of hallucinations as the response length grows. Only the average number of hallucinations in news summarization tasks significantly increases with the length of the context. This may be because the contexts in the other two tasks are more structured, and an increase in length does not significantly raise the difficulty of understanding the content.

\begin{figure}[t]
    \includegraphics[width=0.49\textwidth]{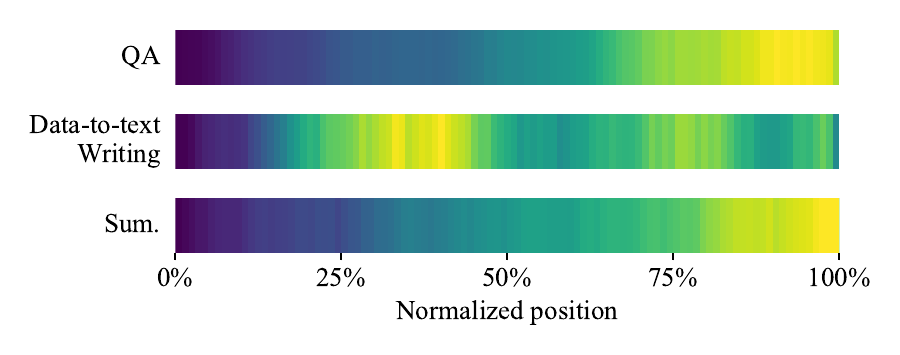}
    \caption{Heatmaps of normalized hallucination occurrence positions.
    The probability of hallucinations occurring is higher in brighter areas.
    }
    \label{fig:halu_pos_heatmap}
\end{figure}

\paragraph{Location of Hallucinations} In Figure~\ref{fig:halu_pos_heatmap}, we present the heatmap of the hallucination occurrence positions. Hallucinations are significantly more likely to occur towards the end of responses in question-answering and news summarization tasks. Compared to other tasks, the data-to-text writing task has a relatively higher occurrence of hallucinations in the first half.
In that bright area, hallucinations concerning business attributes frequently occur.

\section{Experimental Setup}

\subsection{Hallucination Detection Algorithms}



Using {\CorpusName}, we conducted experiments with the following four distinct algorithms for hallucination detection:

\noindent \textbf{Hallucination Detection Prompt}: Hallucination detection prompts are manually crafted to instruct LLMs~(GPT-4-turbo and GPT-3.5-turbo) in assessing whether a given reference-response pair contains hallucinated content and to identify the corresponding hallucinated spans in the response. 
For detailed information about these prompts, please refer to Appendix~\ref{sec:detect_prompts}. 

\begin{table*}[t]
\scriptsize
\centering
    \begin{tabular}{l|ccccccccc|ccc}
    \toprule
    \multirow{2}{*}{Methods}
     & \multicolumn{3}{c}{\textsc{Question Answering}} & \multicolumn{3}{c}{\textsc{Data-to-text Writing}} & \multicolumn{3}{c}{\textsc{Summarization}} & \multicolumn{3}{c}{\textsc{Overall}}  \\
     \cmidrule(lr){2-4} \cmidrule(lr){5-7} \cmidrule(lr){8-10}  \cmidrule(lr){11-13} 
     & Precision & Recall & F1 & Precision & Recall & F1 & Precision & Recall & F1 & Precision & Recall & F1\\
     \midrule
     Prompt$_\textrm{gpt-3.5-turbo}$ & 18.8 & 84.4 & 30.8 & 65.1 & 95.5 & 77.4 & 23.4 & 89.2 & 37.1 & 37.1 & 92.3 & 52.9\\
     Prompt$_\textrm{gpt-4-turbo}$ & 33.2 & 90.6 & 45.6 & 64.3 & \textbf{100.0} & 78.3 & 31.5 & 97.6 & 47.6 & 46.9 & 97.9 & 63.4\\
     SelfCheckGPT$_\textrm{gpt-3.5-turbo}$ & 35.0 & 58.0 & 43.7 & 68.2 & 82.8 & 74.8 & 31.1  & 56.5 & 40.1 & 49.7 & 71.9 & 58.8\\
     LMvLM$_\textrm{gpt-4-turbo}$ & 18.7 & 76.9 & 30.1 &68.0 & 76.7 & 72.1 & 23.3 & 81.9 & 36.2 & 36.2 & 77.8 & 49.4\\
     Finetuned Llama-2-13B & \textbf{61.6} & 76.3 & \textbf{68.2} & \textbf{85.4} & 91.0 &\textbf{88.1} & \textbf{64.0} & 54.9 &\textbf{59.1} &\textbf{76.9} & 80.7 & \textbf{78.7}  \\
    \bottomrule
    \end{tabular}
    \caption{The response-level hallucination detection performance for each baseline method across different tasks and different models. 
    }
    \label{tab:5-detection-result}
\end{table*}

\begin{table*}[!ht]
\scriptsize
\centering
    \begin{tabular}{l|ccccccccc|ccc}
    \toprule
    \multirow{2}{*}{Methods} & \multicolumn{3}{c}{\textsc{Question Answering}} & \multicolumn{3}{c}{\textsc{Data-to-text Writing}} & \multicolumn{3}{c}{\textsc{Summarization}} & \multicolumn{3}{c}{\textsc{Overall}}\\
     \cmidrule(lr){2-4} \cmidrule(lr){5-7} \cmidrule(lr){8-10}  \cmidrule(lr){11-13}
     & Precision & Recall & F1  & Precision & Recall & F1  & Precision & Recall & F1 & Precision & Recall & F1 \\
     \midrule
     Prompt Baseline$_\textrm{gpt-3.5-turbo}$ & 7.9 & 25.1 & 12.1 & 8.7 & 45.1 & 14.6 & 6.1 & 33.7  & 10.3 & 7.8 & 35.3 & 12.8 \\
     Prompt Baseline$_\textrm{gpt-4-turbo}$& 23.7& 52.0 & 32.6  & 17.9 & \textbf{66.4} & 28.2 & 14.7 & \textbf{65.4} & 24.1 & 18.4 & \textbf{60.9} & 28.3\\
     Finetuned Llama-2-13B & \textbf{55.8} & \textbf{60.8} & \textbf{58.2} & \textbf{56.5} & 50.7 & \textbf{53.5} & \textbf{52.4} & 30.8 & \textbf{38.8} & \textbf{55.6} & 50.2 & \textbf{52.7}\\
    \bottomrule
    \end{tabular}
    \caption{The span-level detection performance for each baseline method across different tasks and different models. 
    %
    }
    \label{tab:2-span-result}
\end{table*}

\noindent \textbf{SelfCheckGPT}~\citep{Manakul2023-selfcheckgpt}: SelfCheckGPT employs a zero-resource, sampling-based method to fact-check the responses of black-box models. When processing each response in {\CorpusName}, 3 extra responses from the same model were sampled and served as references, and GPT-3.5-turbo was used to verify consistency. We detected hallucinations sentence-by-sentence within a response, and then aggregated these results to provide a response-level detection outcome.

\noindent \textbf{LMvLM}~\citep{cohen2023lm}: LMvLM is an approach that employs a multi-turn interaction between two Language Models that aim to discover inconsistencies through cross-examination.

\noindent \textbf{LLM Finetuning}: Llama-2-13B has been fine-tuned using the training set from {\CorpusName}. The model takes the context-response pair with proper instructions as the input and treats the hallucinate span as the targeted generation output. We employed full training  with an initial learning rate of 2e-5, and limiting the training to 1 epochs, all conducted on 4 A100 GPUs. 

\subsection{Data Split}
All detection algorithms are tested on the same {\CorpusName} test set, which consists of 450 instances in total, derived by randomly selecting 150 instances from each task type. The rest of the data is used to fine-tune the LLama-2-13B model, as previously mentioned.

\subsection{Evaluation Metrics}
It is a more challenging and significant task to identify the locations of hallucinations within the response than only determining whether a response contains hallucinations. We assess hallucination detection at both the response and span levels.

\paragraph{Response-level Detection} We report precision, recall, and F1 score for each detection algorithm and its variants across different tasks.
\paragraph{Span-level Detection} We calculate the overlap between the detected span and human-labeled span and report the precision, recall, and f1 score at the char-level.

\section{Experimental Results} 

\subsection{Response-level Detection}
The results in Table~\ref{tab:5-detection-result} reveal that hallucination detection remains a significant challenge in the context of RAG for all existing detection methods. Even when reference information is available, the responses generated may still include hallucinations, which current LLMs cannot reliably identify. The most advanced LLM, GPT-4-turbo, achieves only an average F1 score of 63.4\%. For another notable baseline, SelfCheckGPT also shows unsatisfactory performance in this regard, achieving an average F1 score of 58.8\% with GPT-3.5-turbo.

By utilizing our high-quality training set, a fine-tuned Llama-2-13B can achieve the best performance with an average 78.7\% f1 score. This shows the effectiveness of our data in improving the model's hallucination detection ability.

\subsection{Span-level Detection}
{\CorpusName}, as a hallucination corpus with fine-grained span labels, enables us to present experimental results for span-level detection, serving as a baseline for future research. As shown in Table~\ref{tab:2-span-result}, the overall performance of the current detection method is sub-optimal, highlighting the challenges in span-level detection. Even the advanced GPT-4-turbo tends to incorrectly classify many non-hallucinated contents with a low precision of 18.4\%. While our fine-tuned model shows improved capability in identifying hallucinated spans by achieving an averaged f1 score of 52.7\%, it still falls short of perfect detection, emphasizing the inherent difficulties of this task.

\begin{table*}[!t]
\scriptsize
\centering
    \begin{tabular}{c|c|cc}
    \toprule
    \textsc{Group} & \textsc{Selection Strategy} & \textsc{Valid Response Num} & \textsc{Hallucination Rate} \\
     \midrule
     \multirow{3}{*}{{\makecell{Llama-2-7B-chat~(51.8)\\Mistral-7B-Instruct~(57.6)}}} &  Random & 450  & 52.4(-) \\
       & Select the response with fewer detected hallucination spans & 450 & 41.1(\textbf{$\downarrow$21.6\%}) \\
       & Select the response with no detected hallucination spans & 328$^{\dagger}$  & 19.3(\textbf{$\downarrow$63.2\%}) \\
     \midrule\midrule
     \multirow{3}{*}{\makecell{GPT-3.5-Turbo-0613~(10.9)\\GPT-4-0613~(9.3)}} &  Random & 450  & 9.8(-) \\
     & Select the response with fewer detected hallucination spans & 450  & 5.6(\textbf{$\downarrow$42.9\%}) \\
     & Select the response with no detected hallucination spans & 448$^{\dagger}$  & 4.8(\textbf{$\downarrow$51.0\%}) \\
    \bottomrule
    \end{tabular}
    \caption{Utilizing the finetuned hallucination detector to sample from two responses can significantly reduce the rate of hallucinations. The numbers within the brackets in the group column represent the model's hallucination rate. ${\dagger}$: Some instances did not have responses that met the required criteria.}
    \label{tab:7-suppression}
\end{table*}

We also report the detection performance across four different types of hallucination spans. In the current stage, as we have not differentiated the types of detected hallucinations, we only report the char-level recall for different types of hallucinations. As indicated in Figure~\ref{fig:type_recall}, the detection of evident hallucinations proves more effective compared to that of subtle hallucinations.

\begin{figure}[t]
    \includegraphics[width=0.5\textwidth]{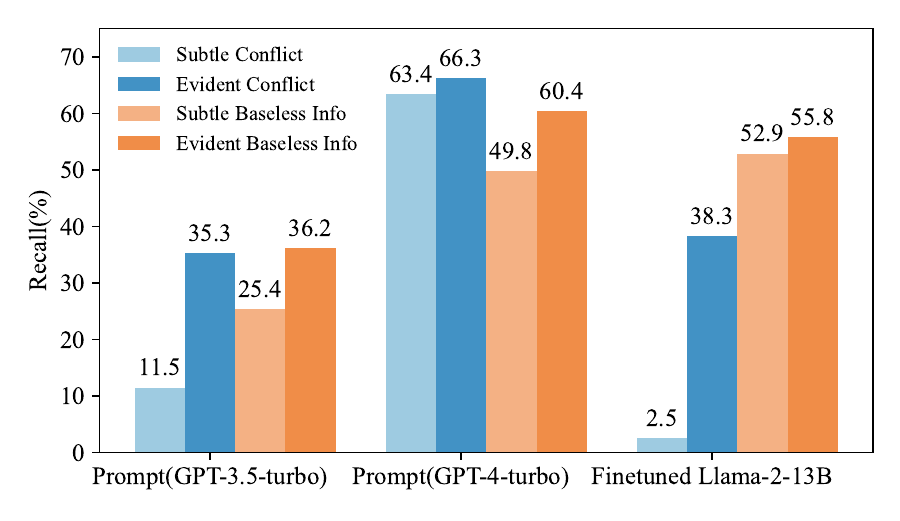}
    \caption{The span-level recalls of different models on four types of hallucinations.}
    \label{fig:type_recall}
\end{figure}

\subsection{Hallucination Suppression}
We tested the effectiveness of hallucination suppression using our finetuned hallucination detection model. For the 450 instances in the test set, we employed two strategies to select a final output from two responses generated by two different models with similar hallucination densities. The first strategy involved selecting the response with fewer predicted hallucination spans. The second strategy, more stringent, mandated that the selected response have no detected hallucination spans. When the number of hallucination spans detected in both candidate responses is the same, one will be chosen at random. Due to limited response candidates, not all instances have a response that conforms to the second strategy. In practical scenarios, this issue can be addressed by increasing the number of candidate responses. We employed random selection as a simple baseline for comparison.

The results shown in Table~\ref{tab:7-suppression} indicate that with the help of the hallucination detector, both strategies can significantly reduce the hallucination rate. For the relatively small Llama-2-7B-chat and Mistral-7B-Instruct models, compared to random selection, the first strategy reduced the hallucination rate by 21.6\%, while the second strategy achieved a reduction of 63.2\%. Even for models with a low hallucination rate, specifically GPT-3.5-Turbo and GPT-4, employing the finetuned hallucination detector for sampling can still further reduce the rate of hallucinations. The two strategies yielded a reduction in hallucination rates of 42.9\% and 51.0\%, respectively. These results demonstrate the potential of an efficient hallucination detection model in developing trustworthy RAG LLMs.

\section{Conclusion}
In this paper, we introduce {\CorpusName}, a large-scale corpus of naturally generated hallucinations, featuring detailed word-level annotations tailored for RAG scenarios. Our work includes an in-depth analysis of the interplay between hallucinations and various factors, such as task types, models being used, and contextual settings. 

Additionally, we conduct empirical benchmarks of several hallucination detection approaches using our corpus. We show that fine-tuning Llama with {\CorpusName} leads to competitive performance. This implies that by using a high-quality dataset such as {\CorpusName}, it is possible to develop specialized hallucination detection models that are highly effective when compared to prompt-based methods using general models such as GPT-4. 

Simultaneously, our findings reveal that identifying hallucinations in RAG contexts, particularly at the span level, remains a formidable challenge, with current methods still falling short of reliable detection. 
We hope that {\CorpusName}, 
can assist the development of hallucination detection techniques for retrieval augmented generation. 


\section{Limitations}

The study of hallucination in large language models is a rapidly advancing field, characterized by the continuous evolution of application scenarios, sources of hallucination, and techniques for detecting and preventing them. While our work represents the first attempt to benchmark hallucination in the RAG setting, there may be situations not addressed by this research that are nonetheless significant for certain practical applications.

\section{Ethical considerations}
This work is in full compliance with the Ethics Policy of the ACL. 
We acknowledge that responses generated by LLMs in this study may contain inaccuracies. 
Aside from this, to the best of our knowledge, there are no additional ethical issues associated with this paper.

\section{Acknowledgement}
We appreciate the valuable feedback and assistance from Shizhe Diao. We thank Doris Li for her support in creating the illustrations for this research.

\bibliography{anthology,custom}
\bibliographystyle{acl_natbib}

\appendix

\onecolumn
\clearpage
\section{Hallucination Examples}
\label{sec:example}
\begin{table*}[!ht]

\renewcommand{\arraystretch}{1.1}

\small
\centering
    \begin{tabular}{p{1.0\textwidth}}
    \toprule
    \textbf{\textsc{Evident Conflict}}\\
    \midrule
    \textbf{Example 1}\\
    Original content: They’ve sent about 500 of the most seriously injured people for treatment in Kathmandu already -- the majority by road.\\
    Generative content: The team airlifted 500 injured people… to Kathmandu for medical treatment. \\
    \\
    \textbf{Example 2}\\
    Original content: "RestaurantsReservations": false\\
    Generative content: The restaurant offers outdoor seating and takes reservations\\
    \midrule
    \textbf{\textsc{Subtle Conflict}}\\
    \midrule
    \textbf{Example 1}\\
    Original content: This is the section that lets the government deny trademark protection to a mark that is "immoral," "scandalous" or "disparaging."\\
    Generative content: …violates Section 2(a) of the trademark act, which prohibits marks that are "immoral, scandalous, or disqualified."\\
    \\
    \textbf{Example 2}\\
    Original content: It was quite a hassle to find a table since this place is crowded all the time, but well, nothing can stand between me and that absolutely delightful lobster roll\\
    Generative content: The restaurant is located in a crowded area, but the delicious lobster rolls make it worth the hassle to find a table.\\
    \midrule
    \textbf{\textsc{Evident Introduction of Baseless Information}} \\
    \midrule
    \textbf{Example 1}\\
    Original content: The couple divorced in 1968… She had no contact with the surviving members of the band until meeting up at the 2006 Las Vegas premiere of "The Beatles Love." \\
    Generative content: The couple divorced in 1968, but later reconciled and attended the 2006 Las Vegas premiere of "The Beatles Love." \\
    \\
    \textbf{Example 2}\\
    Original content: “RestaurantsTakeOut”: true\\
    Generative content: Customers can also enjoy takeout and delivery options.\\
    \midrule
    \textbf{\textsc{Subtle Introduction of Baseless Information}} \\
    \midrule
    \textbf{Example 1}\\
    Original content: Amazing happy hour and great people.\\
    Generative content: One of the standout features of Killer Shrimp is their happy hour, which is available daily from 4 pm to 7 pm.\\
    \\
    \textbf{Example 2}\\
    Original content: Place mashed potatoes in a medium bowl. Stir in onions, cheese, salt, pepper, and bacon.\\
    Generative content: Place your mashed potatoes in a medium bowl. Stir in cooked onions, cheese, salt, pepper, and bacon.\\
    \bottomrule
    \end{tabular}
    \caption{Examples of the four types of hallucinations.} 
    \label{tab:examples}
\end{table*}

\clearpage

\section{Response Generation Prompts}
\label{sec:prompts}
\begin{table*}[!ht]

\renewcommand{\arraystretch}{1.1}

\small
\centering
    \begin{tabular}{p{1.0\textwidth}}
    \toprule

    \textbf{\textsc{Question answering}}\\
    \midrule
    Answer the following question:\\
    \{question\}\\
    Bear in mind that your response should be strictly based on the following 3 passages:\\
    \{passages\}\\
    In case the passages do not contain the necessary information to answer the question, please reply with: "Unable to answer based on given passages."\\
    \midrule
    \textbf{\textsc{Data-to-text Writing}}\\
    \midrule
    Instruction:\\
    Write an objective overview about the following local business based only on the provided structured data in the JSON format.\\
    You should include details and cover the information mentioned in the customers' review. The overview should be 100 - 200 words. Don't make up information.\\
    Structured data:\\
    \{json\_data\}\\
    Overview:\\
    \midrule
    \textbf{\textsc{Summarization}}\\
    \midrule
    Summarize the following news within \{word\_num\} words:\\
    \{news\}\\
    output:\\
    \bottomrule
    \end{tabular}
    \caption{Prompts for generating responses for the three types of tasks. word\_num is min(200, word\_num\_of\_news//4). The word count requirement is only to control the length of the generated summarization, it will not serve as the basis for hallucination annotation.} 
    \label{tab:prompts}
\end{table*}

\clearpage
\section{Annotation Details}
\label{sec:interface}
\begin{figure*}[h]
    \includegraphics[width=0.98\textwidth]{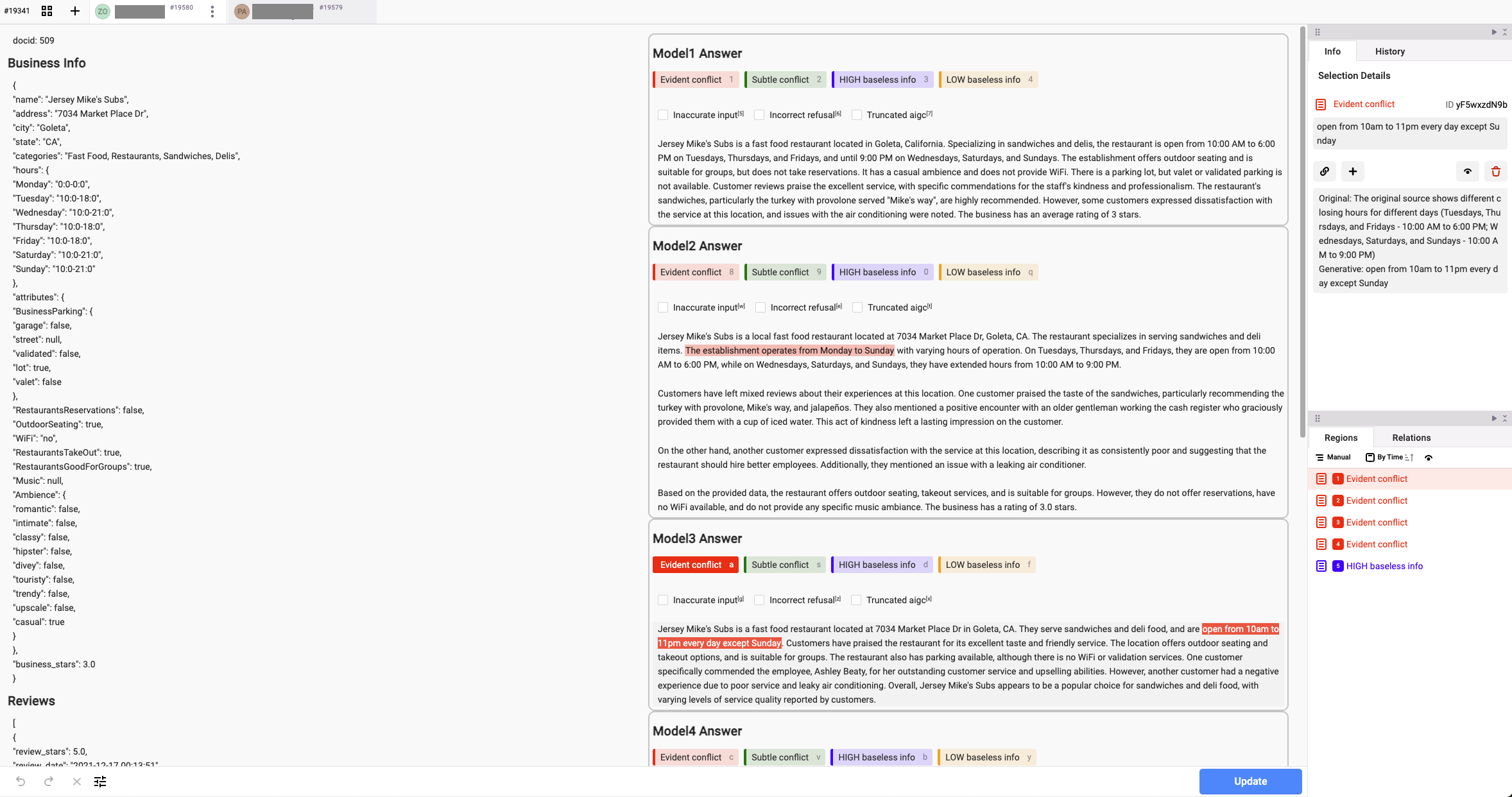}
    \caption{Annotation interface. For privacy reasons, we have masked the full names of the annotators in the screenshot.}
    \label{fig:type_recall}
\end{figure*}


\begin{table*}[h]
\centering
\small
\begin{tabular}{c|cccccc}
\toprule
\multirow{2}{*}{Task}                 & \multirow{2}{*}{Model} & \multirow{2}{*}{\# Hallucination Span} & \multicolumn{2}{c}{\it{implicit\_true}} & \multicolumn{2}{c}{\it{due\_to\_null}} \\
                                      &                        &                                        & \# Span          & \% Span         & \# Span         & \% Span         \\
\midrule
\multirow{6}{*}{Question Answering}   & GPT-3.5-turbo-0613     & 89                                     & 33               & 0.371           &                 &                 \\
                                      & GPT-4-0613             & 51                                     & 15               & 0.294           &                 &                 \\
                                      & Llama-2-7B-chat        & 1010                                   & 251              & 0.249           &                 &                 \\
                                      & Llama-2-13B-chat       & 654                                    & 215              & 0.329           &                 &                 \\
                                      & Llama-2-70B-chat       & 529                                    & 168              & 0.318           &                 &                 \\
                                      & Mistral-7B-Instruct    & 594                                    & 164              & 0.276           &                 &                 \\
\midrule
\multirow{6}{*}{Data-to-text Writing} & GPT-3.5-turbo-0613     & 384                                    & 52               & 0.135           & 69              & 0.180           \\
                                      & GPT-4-0613             & 354                                    & 24               & 0.068           & 209             & 0.590           \\
                                      & Llama-2-7B-chat        & 1775                                   & 195              & 0.110           & 230             & 0.130           \\
                                      & Llama-2-13B-chat       & 2803                                   & 260              & 0.09            & 439             & 0.157           \\
                                      & Llama-2-70B-chat       & 1834                                   & 274              & 0.149           & 272             & 0.148           \\
                                      & Mistral-7B-Instruct    & 2140                                   & 102              & 0.048           & 423             & 0.198           \\
\midrule
\multirow{6}{*}{Summarization}        & GPT-3.5-turbo-0613     & 60                                     & 14               & 0.233           &                 &                 \\
                                      & GPT-4-0613             & 80                                     & 10               & 0.125           &                 &                 \\
                                      & Llama-2-7B-chat        & 517                                    & 44               & 0.085           &                 &                 \\
                                      & Llama-2-13B-chat       & 342                                    & 28               & 0.082           &                 &                 \\
                                      & Llama-2-70B-chat       & 245                                    & 27               & 0.110           &                 &                 \\
                                      & Mistral-7B-Instruct    & 828                                    & 52               & 0.063           &                 &                 \\
\midrule
Overall                               &                        & 14289                                  & 1928             & 0.135           & 1642                                                  & 0.115 \\
\bottomrule
\end{tabular}
\caption{Detailed statistical information for the labels \textit{implicit\_true} and \textit{due\_to\_null}. 
The majority of implicit truths appear in two types of tasks: question answering and data-to-text writing.
About 17.7\% hallucination spans in the data-to-text writing tasks are related to null values in the JSON data.}
\label{tab:implicit_true}
\end{table*}
\clearpage

\section{Hallucination Detection Prompts}
\label{sec:detect_prompts}
\begin{table*}[!ht]

\renewcommand{\arraystretch}{1.1}

\small
\centering
    \begin{tabular}{p{1.0\textwidth}}
    \toprule
    \textbf{\textsc{Summarization}}\\
    \midrule
    Below is the original news:\\
\{article\}\\
Below is a summary of the news:\\
\{summary\}\\
Your task is to determine whether the summary contains either or both of the following two types of hallucinations:\\
1. conflict: instances where the summary presents direct contraction or opposition to the original news;\\
2. baseless info: instances where the generated summary includes information which is not substantiated by or inferred from the original news. \\
Then, compile the labeled hallucinated spans into a JSON dict, with a key "hallucination list" and its value is a list of hallucinated spans. If there exist potential hallucinations, the output should be in the following JSON format: \{"hallucination list": [hallucination span1, hallucination span2, ...]\}. Otherwise, leave the value as a empty list as following: \{"hallucination list": []\}.\\
Output:\\
    \midrule
    \textbf{\textsc{Question answering}}\\
    \midrule
        Below is a question:\\
\{question\}\\
Below are related passages:\\
\{passages\}\\
Below is an answer:\\
\{answer\}\\
Your task is to determine whether the answer contains either or both of the following two types of hallucinations: \\
1. conflict: instances where the answer presents direct contraction or opposition to the passages;\\
2. baseless info: instances where the answer includes information which is not substantiated by or inferred from the passages.\\
Then, compile the labeled hallucinated spans into a JSON dict, with a key "hallucination list" and its value is a list of hallucinated spans. If there exist potential hallucinations, the output should be in the following JSON format: \{"hallucination list": [hallucination span1, hallucination span2, ...]\}. Otherwise, leave the value as a empty list as following: \{"hallucination list": []\}.\\
Output:\\
    
    \midrule
    \textbf{\textsc{Data-to-text Writing}}\\
    \midrule
Below is a structured data in the JSON format:\\
\{business info\}\\
Below is an overview article written in accordance with the structured data:\\
\{overview\}\\
Your task is to determine whether the overview contains either or both of the following two types of hallucinations: \\
1. conflict: instances where the overview presents direct contraction or opposition to the structured data;\\
2. baseless info: instances where the generated overview includes information which is not substantiated by or inferred from the structured data.\\
In JSON, "null" or "None" represents an unknown value rather than a negation. \\
Then, compile the labeled hallucinated spans into a JSON dict, with a key "hallucination list" and its value is a list of hallucinated spans. If there exist potential hallucinations, the output should be in the following JSON format: \{"hallucination list": [hallucination span1, hallucination span2, ...]\}. Otherwise, leave the value as a empty list as following: \{"hallucination list": []\}.\\
Output:\\
    \bottomrule
    \end{tabular}
    \caption{Prompts for detecting hallucination for the three types of tasks. In the prompt for data-to-text writing, we clarified that null or None in JSON should be treated as unknown rather than a negation.} 
    \label{tab:prompts}
\end{table*}

\end{document}